# BARD10: A New Benchmark Reveals Significance of Bangla Stop-Words in Authorship Attribution

Authors:

1. Abdullah Muhammad Moosa, Graduate, Department of Mechatronics & Industrial Engineering, Chittagong University of Engineering & Technology, Chittagong 4349, Bangladesh.

abdullahmoosa128@gmail.com

2. Nusrat Sultana, Lecturer, Department of Mechatronics & Industrial Engineering, Chittagong University of Engineering & Technology, Chittagong 4349, Bangladesh.

nusratsultana@cuet.ac.bd

3. Mahdi Muhammad Moosa, Assistant Professor, Department of Mathematics & Natural Sciences, Brac University , Dhaka 1212, Bangladesh.

mahdi.moosa@bracu.ac.bd

4. Md. Miraiz Hossain, Assistant Professor, Department of Mechatronics & Industrial Engineering, Chittagong University of Engineering & Technology, Chittagong 4349, Bangladesh.

miraizhossain@cuet.ac.bd

## Abstract

This research presents a comprehensive investigation into Bangla authorship attribution, introducing a new balanced benchmark corpus BARD10 (Bangla Authorship Recognition Dataset of 10 authors) and systematically analyzing the impact of stop-word removal across classical and deep learning models to uncover the stylistic significance of Bangla stop-words. BARD10 is a curated corpus of Bangla blog and opinion prose from ten contemporary authors, alongside the methodical assessment of four representative classifiers: SVM (Support Vector

Machine), Bangla BERT (Bidirectional Encoder Representations from Transformers), XGBoost, and a MLP (Multilayer Perception), utilizing uniform preprocessing on both BARD10 and the benchmark corpora BAAD16 (Bangla Authorship Attribution Dataset of 16 authors). In all datasets, the classical TF-IDF + SVM baseline outperformed, attaining a macro-F1 score of 0.997 on BAAD16 and 0.921 on BARD10, while Bangla BERT lagged by as much as five points. This study reveals that BARD10 authors are highly sensitive to stop-word pruning, while BAAD16 authors remain comparatively robust highlighting genre-dependent reliance on stop-word signatures. Error analysis revealed that high frequency components transmit authorial signatures that are diminished or reduced by transformer models. Three insights are identified: Bangla stop-words serve as essential stylistic indicators; finely calibrated ML models prove effective within short-text limitations; and BARD10 connects formal literature with contemporary web dialogue, offering a reproducible benchmark for future long-context or domain-adapted transformers.

# 1 | INTRODUCTION

The growing volume of Bangla text on online platforms has created a strong need for reliable authorship identification. Authorship attribution aims to determine the most likely author of a text by analyzing linguistic and stylistic patterns, and it has proven valuable in plagiarism detection, literary analysis, and digital forensics [1]. Most prior studies have focused on high-resource languages such as English, where large and balanced datasets support model development [2]. In contrast, Bangla, although the seventh most spoken language globally has received limited computational attention. Its complex morphology, agglutinative structure, and flexible word order make conventional stylometric techniques less reliable [3], [4]. Recent advances in stylometry have explored both classical feature-based models and deep architectures. For instance, hybrid models using syntactic and lexical features continue to perform competitively in authorship detection tasks [5]. However, in Bangla, the lack of standardized corpora and consistent preprocessing practices still hinders fair comparison and reproducibility.

This research addresses these gaps by introducing BARD10, a balanced dataset containing texts from ten contemporary Bangla authors. We systematically evaluate both traditional and neural models SVM, XGBoost, MLP, and Bangla BERT under identical preprocessing pipelines, focusing on the impact of stop-word removal. The results show that Bangla stop-words often act as stylistic indicators and should not always be removed during text cleaning. By offering a new benchmark corpus, reproducible evaluation, and linguistic insights, this study provides a foundation for more transparent and resource-aware research in Bangla authorship attribution.

# 2 | Related Work

Recent research in authorship attribution has shifted toward pretrained language models and hybrid methods that integrate linguistic features with deep contextual embeddings. Huertas-Tato et al. [6] proposed the PART framework, which applies a contrastive learning objective on a pretrained Transformer backbone and demonstrates strong zero-shot attribution performance. Similarly, Huang and Iwaihara [7] explored short-text authorship identification using capsule networks built on top of pretrained models, showing that contextual embeddings outperform traditional hand-crafted features. For less-resourced languages, a hybrid Transformer that combines contextual embeddings with linguistic descriptors achieved robust accuracy on Romanian corpora [8], suggesting the suitability of such models when balanced datasets and linguistic resources are limited.

In Bangla, early computational studies focused on machine learning–based stylometric modeling. Phani et al. [9] applied several statistical and supervised learning algorithms for authorship attribution in Bangla blogs, revealing that lexical and stylistic cues can effectively characterize authorial style across informal text domains. Later, Islam et al. [10] introduced a neural stylometric framework that integrates traditional handcrafted features with feed-forward neural networks for Bangla literature, achieving improved accuracy over earlier SVM-based approaches. These works established the foundation for feature-driven authorship identification in Bangla, yet they did not explore deep contextual embeddings or examine how preprocessing choices, such as stop-word removal, influence classifier performance. Subsequent research has extended these efforts using deep learning and multilingual hybrid frameworks. Anwar et al.[11] introduced an optimized ML + DL framework combining TF-IDF features with LSTM and GRU networks on the Reuters 50_50 dataset, achieving higher accuracy than classical algorithms through hyperparameter tuning and feature selection. Likewise, Tang [12] proposed a CNN–LSTM model with an attention mechanism for literary authorship attribution in Chinese texts, demonstrating that multi-level feature extraction enhances recognition robustness across genres. Imran and Amin [3] fine-tuned transformer architectures such as bnBERT and mBERT on the BAAD16 dataset, obtaining competitive results yet without examining preprocessing sensitivity. The BAAD16 corpus itself [13] provided an important step toward balanced data but remains limited to canonical literary text. More recently, Khatun et al. [4] applied transfer learning using ULMFiT to Bangla literature, though their work does not assess the influence of stop-word handling or feature ablation. Haque et al. [14] addressed sentiment classification in Bangla social-media text using machine learning and BNLP preprocessing tools, offering insights into modern tokenisation and stop-word management relevant to authorship tasks. At the broader NLP level, survey work by Huang et al. [15] on authorship attribution in the era of large language models highlights persistent challenges in cross-domain generalisation, explainability, and dataset imbalance.

From this body of work, three major research gaps emerge. First, existing studies on Bangla authorship attribution rarely conduct systematic comparisons of preprocessing strategies, such as stop-word removal across both classical machine learning models and deep learning architectures. Second, there remains a scarcity of publicly available, balanced benchmark corpora that enable fair comparison between traditional algorithms like SVM or XGBoost and Transformer-based approaches under consistent preprocessing conditions. Third, while many works emphasize improving model performance, few investigate how stylistic features such as stop-words or stop-words influence attribution accuracy, particularly in morphologically rich languages like Bangla. Addressing these limitations, the present study introduces the BARD10 dataset, performs a controlled evaluation of classical and Transformer models, and provides the first empirical analysis of how stop-word removal affects authorship attribution in Bangla texts.

## 3 | Methodology

### 3.1 Corpora

**3.1.1 Bangla Authorship Attribution Dataset (BAAD16)**

The BAAD16 [13] collection includes literary works from 16 notable Bangla authors (fiction genre). The overall word counts for each author differ significantly, spanning from around 2.7 million words (Humayun Ahmed) to about 111,000 words (Zahir Rayhan). The dataset comprises renowned individuals such Rabindranath Tagore, Sarat Chandra Chattopadhyay, and Satyajit Ray, among others. Normalization of Unicode characters and elimination of all punctuation were executed during preprocessing. No supplementary filtering or deduplication was implemented. The dataset was employed to examine stylistic variance among prominent Bangla literary luminaries. Table 1 indicates that, among the sixteen authors, stop-words comprise an average of 15.2% of the vocabulary, with individual proportions closely grouped between around 14% and 17%. This tight range indicates a generally similar dependence on stop-words; yet the minor inter-author variations may still provide valuable stylistic indicators for authorship modelling.

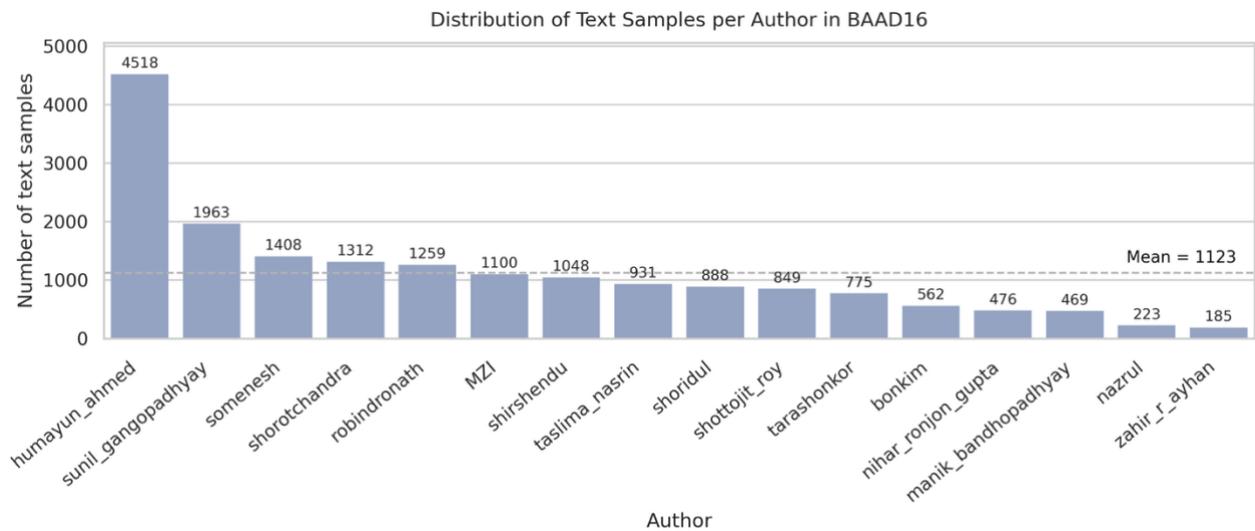

**Figure 1:** Distribution of text samples per author in the BAAD16 corpus.

The bar heights in Figure 1 represent the total number of 750-word segments provided by each of the sixteen canonical Bangla writers, while the dashed line is the corpus-wide average of 1,123 samples. It demonstrates a significantly imbalanced class distribution: Humayun Ahmed alone provides almost 4,500 segments more than four times the corpus average, while the least represented writers provide fewer than 250. This mismatch might skew supervised models in favor of prevalent classes, exaggerate naïve accuracy, and obscure inadequate performance in minority classes. Consequently, subsequent experiments use stratified assessment splits and, where applicable, class-balancing methodologies such as weighted loss, focus loss to alleviate these effects.

**Table 1: Total word tokens, stop-word counts and stop-word ratios for each author in the BAAD16 dataset**

| Author Id | Author | Total_Words | Stopword_Count | Stopword_% |
|---|---|---|---|---|
| A0 | MZI | 1755800 | 273713 | 15.589 |
| A1 | bongkim | 1007573 | 135935 | 13.491 |
| A2 | Humayun Ahmed | 7122642 | 1060318 | 14.887 |
| A3 | Manik Bandopadhyay | 764482 | 111844 | 14.63 |
| A4 | nazrul | 354056 | 59838 | 16.901 |
| A5 | nihar_ronjon_gupta | 791865 | 129439 | 16.346 |
| A6 | Rabindranath Tagore | 2130389 | 329352 | 15.46 |
| A7 | Shirshendu Mukhopadhyay | 1616393 | 243882 | 15.088 |
| A8 | shomresh | 2247086 | 338215 | 15.051 |
| A9 | shordindu | 1608057 | 217453 | 13.523 |
| A10 | shorotchandra | 2212127 | 327353 | 14.798 |
| A11 | shottojit_roy | 1412123 | 223622 | 15.836 |
| A12 | shunil_gongopaddhay | 3199729 | 516800 | 16.151 |
| A13 | tarashonkor | 1323714 | 197614 | 14.929 |
| A14 | toslima_nasrin | 1524168 | 240221 | 15.761 |
| A15 | zahir_rayhan | 285928 | 42100 | 14.724 |

### 3.1.2 Bangla Authorship Recognition Dataset of 10 authors (BARD10)

A unique dataset, BARD10, was developed for this study to assess authorship attribution in a more modern and casual context. This compilation features works from ten contemporary Bangla authors, predominantly obtained from online blogs, with select examples included in Table 2. The chosen blogs encompass both fiction and opinion based genres. The dataset is equitably distributed among authors, with the quantity of stories varying from few dozen to more than one hundred per author. Average text lengths vary from roughly 150 to more than 1400 words. During collecting, texts were sanitized by Unicode normalization and the elimination of punctuation, including indicators peculiar to Bangla. This dataset establishes a novel baseline for Bangla authorship authentication in non-canonical literary styles. The

Author Id mentioned in the Table 2 will be used in the following sections to mention respective authors.

**Table 2: Data samples of BARD10 datasets and total word counts and proportion of stop-words for each author.**

| Author ID | Author | Sample_Text | Total_Words | Stopword_Count | Stopword_% |
|---|---|---|---|---|---|
| A0 | Akash Ambar | রাগ ক্রোধ ঘৃণা আবেগ আর অপ্রকাশ্য স্মৃতিভাবখানি বাদ দিলে যা কিছু থাকে তাকে নামিয়ে আনা কিংবা না-আনার প… | 487791 | 76490 | 15.7 |
| A1 | Noyon | অনেকেই প্রিয় গল্পকারদের খুঁজে বেড়ান, তাদের লেখা গল্পগুলো পড়তে চান। অনেকে আবার কবিতা/ছড়ার সমঝদার … | 203457 | 29127 | 14.3 |
| A2 | Hasan Mahbub | প্রচণ্ড ক্ষিধের সময় যদি হাতের কাছেই একটি ভালো রেস্তোরা পাওয়া যায়, এবং সেখানে যদি ভালো মানের মোগলাই প… | 685740 | 108529 | 15.8 |
| A3 | Morubhumi Joldossu | বার-বার পিছিয়ে যাচ্ছিলো ভ্রমণের তারিখ। সেন্টমার্টিন | 285175 | 47066 | 16.5 |
| | | বেড়াতে যাওয়ার আয়োজন ঠিক করে বান্দরবান এমনকি ভারত… | | | |
| A4 | Shunya Aranyak | পরোটা কাহিনী পরের পর্বের জন্য তোলা থাকুক। চতুররা সব জানতে ব্যস্ত কারে কতোটুক সুন্দর দেখা গেলো ৭-৮ দি… | 344716 | 53947 | 15.7 |
| A5 | Charu Mannan | \_\_\_\_\_\_\_\_\_\_তুমি বুঝলে না আমায়, তুমি বুঝলে না আমায়, আতসবাজির মত চঞ্চলতা তোমার, তাড়িত করেছিল যৌবণ !… | 68648 | 10436 | 15.2 |
| A6 | Ahmad Abdul Halim | মূল আরবী : মাহমূদ দারবিশ আরবী থেকে অনুবাদ : আহমাদ আবদুল হালিম ছায়া ছায়া, সে পুরুষ কিংবা নারী নয় বাদা… | 166754 | 25082 | 15 |
| A7 | Akhtar Javed | সামনেই শুকনো পাতায় জ্বলে উঠা আগুন জ্বলছে। গুটিসুটি মেরে কিছু গৃহহীন মানুষ জড়ো - আমিও একটু একটু সরে গ… | 76398 | 11519 | 15.1 |

| A8 | Kamal Uddin | রেল লাইন ধরে পায়ে হেঁটে ঢাকার কমলাপুর স্টেশন থেকে চিটাগাং পর্যন্ত যাওয়ার পরিকল্পনার কথা অনেকেই জানেন।… | 159560 | 26047 | 16.3 |
| --- | --- | --- | --- | --- | --- |
| A9 | Tajerul Islam | এই দেহ-মন থেকে একাকার হয়ে কাছে,দূরে ও সবখানে রই চেয়ে, দেখি চির জাগ্রত সে মানুষ করছে সংগ্রাম যুদ্ধ শে… | 78252 | 13668 | 17.5 |

Figure 2 illustrates the quantity of texts attributed to each of the 10 contemporary Bangla writers; the dashed line is the corpus mean of around 234 samples. Morubhumi Joldossu nearly twice the corpus average, whereas Akhtar Javed and Ahmad Abdul Halim each fall below 200 texts.

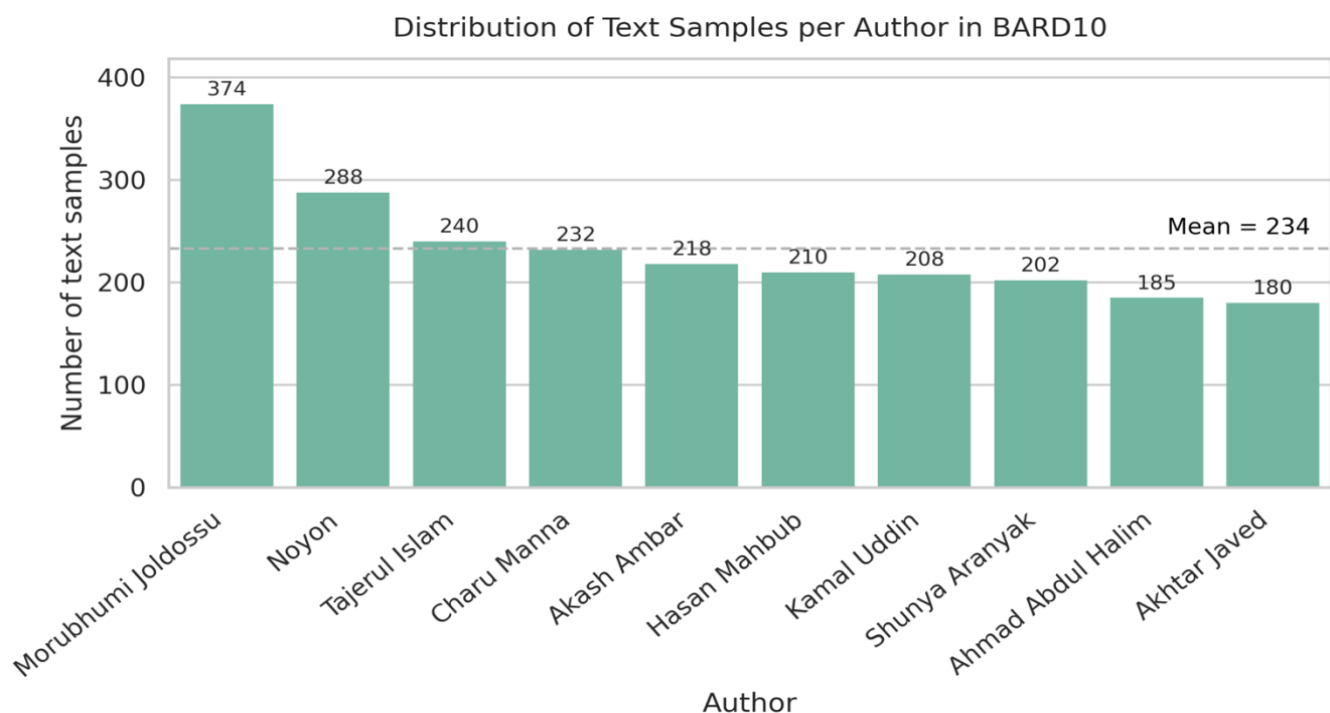

**Figure 2**: Distribution of text samples per author in the BAARD 10 corpus.

Although less severe than BAAD16, this skew may still influence model bias and evaluation equity; hence, stratified splits and loss re-weighting were employed as necessary. In the corpus, stop-words constitute approximately one-sixth of each author's lexicon, exhibiting a little variance of 14.3% to 17.5%. Noyon's writings display the lowest stop-word density at 14.3%, whereas Tajerul Islam demonstrates the greatest at 17.5%, signifying a greater dependence on connective and grammatical indicators. Most authors congregate closely around the sample mean of approximately 15.8%, indicating generally similar style inclinations in their use of stop-words. These data indicate that, despite significant enlargement of the stop-word list, inter-author variations remain minor yet may be indicative for authorship or stylistic analysis.

## 3.2 Experimental Setup

A stratified 80:20 train-test split was created once per dataset (BAAD16 and BARD10) to preserve author proportions and was reused across all experiments. For each dataset, two preprocessing variants were analysed: (i) stop-words retained and (ii) stop-words removed. The training and evaluation procedures, models, and hyper-parameters were held constant across variants; only the pre-processing step was altered to remove stop-words in the non-stop-word condition.

All experiments were executed in Python 3.12 on Linux with a Tesla P100 GPU. Transformer fine-tuning was carried out with PyTorch and HuggingFace Transformers; the TF–IDF, SVM, and metric computations were performed with scikit-learn; gradient boosting was implemented with XGBoost. Bangla-specific preprocessing (text cleaning and stop-word lists) was conducted using BNLP [16]. The embedding-based model was trained with TensorFlow/Keras. Data handling used pandas and NumPy, visualisation employed matplotlib/seaborn, experiment tracking was performed with Weights & Biases (W&B).

## 3.3 Preprocessing

**3.3.1 Punctuation Removal**

Through the use of regular expressions, every single punctuation character, even those that are unique to Bangla, was eliminated. Both sets of data were subjected to this in a consistent manner.

### 3.3.2 Stop Word Removal

The BNLP library [16] provided a standard Bangla stopword list, which was utilised in this study. In the study, this phase was examined as a variable in order to determine the impact that it has on the performance of the model. To prevent leakage from morphological variants, the list was first converted into model-ready tokens by applying the same tokenisation used within each pipeline, the BNLP word tokenizer for neural models and the scikit-learn analyser for TF-IDF. For TF-IDF features, removal was performed after vectorisation: columns whose tokens matched this set were set to zero, leaving all other statistics unchanged. For sequence models, removal was performed at the sequence level: in BERT, stop-word tokens were stripped from the text prior to WordPiece tokenisation; in the MLP, token IDs corresponding to the stop-word set were masked to the padding ID. Through this procedure, both surface forms (e.g., "মতো", "আমি", "অনেক") and the corresponding analyser stems (e.g., "মত", "আম", "অন") were excluded from the model inputs.

### 3.3.3 Additional Preprocessing

Unicode normalization was executed on all texts. Lowercasing was not implemented and stemming or lemmatization were not utilized due to the absence of developed resources in Bangla. Text tokenization for token-based models was executed via TensorFlow's Tokenizer, featuring a vocabulary capacity of 500,000 and an out-of-vocabulary (OOV) token.

## 3.4 Model Training

### 3.4.1 Bangla BERT (Transformer fine-tuning)

The sagorsarker/bangla-bert-base [17] encoder was fine tuned for single sentence classification. Tokenization was carried out with the model's WordPiece tokenizer (max_length = 512, truncation and padding enabled). The classifier head was instantiated as a linear layer over the [CLS] representation with output dimensionality equal to the number of authors (K = 10 for BARD10; K = 16 for BAAD16). Training was executed via the HuggingFace Trainer with epochs = 10, batch size = 16, learning rate = 1e-5, and Adam. Evaluation was performed at each epoch; load_best_model_at_end=True with accuracy as the selection metric was used. The final model was assessed on the held-out test set, and confusion matrices were generated and logged.

### 3.4.2 Linear SVM with TF–IDF

Texts were vectorized using scikit-learn TF–IDF (sublinear_tf=True, use_idf=True). A linear-kernel SVM (sklearn.svm.SVC(kernel="linear", probability=True)) was trained on the TF–IDF matrix. Probability estimates were enabled to support downstream analyses. Test set performance was computed and the confusion matrix was logged.

### 3.4.3 XGBoost with TF–IDF

Nonlinear ensembling over the same TF–IDF features was examined using XGBoost with max_depth = 6, learning_rate = 0.1, n_estimators = 100, objective = "multi:softmax", and random_state = 42. Predictions on the test set were used to compute the evaluation metrics, and confusion matrices were logged to W&B.

### 3.4.4 Embedding based MLP

A fixed Bangla word embedding matrix was constructed from a pretrained Word2Vec model (embedding_dim = 300). Vocabulary induction was performed with a Keras Tokenizer fitted on the training corpus; the top 500,000 indices (max_words = 500000) were allocated, and an embedding matrix of shape (500000, 300) was built. Out-of-vocabulary (OOV) coverage was monitored by counting tokens without a pretrained vector during matrix construction. Inputs were truncated to 4,000 tokens per document.

The network was composed of an embedding layer initialized with the fixed matrix (trainable=False), a GlobalAveragePooling1D layer over token embeddings, and two fully connected layers of sizes 64 and 32 with ReLU activations, followed by a softmax layer with K units (K equal to the number of authors; 10 for BARD10, 16 for BAAD16). Optimization was performed with Adam at learning rate = 0.001, batch size = 64, and a maximum of 500 epochs. Overfitting was mitigated by applying early stopping on validation accuracy with a patience of 70 epochs, with best weights restored. The final model was saved and logged as a W&B artefact; test set predictions were evaluated with the same metric suite, and confusion matrices were plotted via seaborn.

**3.4.5 Reproducibility and logging**

For each run, the following were logged: (i) dataset metadata (split sizes, class count), (ii) model hyper parameters, (iii) training curves, (iv) best model checkpoints (where applicable), and (v) confusion matrices. Exact configurations (maximum sequence length, batch sizes, learning rates, patience) and random seeds were captured in W&B configs to ensure repeatability.

Model performance was evaluated using accuracy, precision, recall, and F1-score. Confusion matrices were examined to characterise class-specific errors. Additional metrics emitted by the training pipeline (micro-F1, weighted precision, macro recall) were also recorded.

## 3.5 Stop-word Sensitivity via Δ-Recall Ablation

To quantify the marginal contribution of individual stop-words, a post-hoc ablation was performed on the trained TF–IDF + linear SVM in which stop-words were retained during training. For each stop-word token w, the corresponding TF–IDF column in the test matrix was set to zero while all other features and model parameters were held fixed; per-author recall was then recomputed. The recall change was defined as

$\Delta \text{Recall}_{w,a} = \text{Recall}_{\text{baseline},a} - \text{Recall}_{\text{after removing } w,a}$ ……………………………………..eq$^n$.(1)

The procedure was executed in parallel over all stop-words, yielding a matrix (rows = stop-words, columns = authors). Values greater than 0 indicate that removing w hurts author a's recall (informative token), values equal to 0 indicate little effect, and values less than 0 indicate that removing w improves recall. Because a single feature column was perturbed at a time with the model held fixed, the measured Δ values isolate inference time contributions without confounding from retraining or IDF shifts. Summation across authors provided a global importance per stop-word; within column ranking identified the most impactful stop-words per author.

# 4 | Results

## 4.1 Overall Performance

This section presents quantitative results for the four classifiers, Bangla BERT, SVM, XGBoost, and MLP on the BAAD16 and BARD10 corpora, along with an analysis elucidating the necessity of maintaining Bangla stop-words for achieving optimal accuracy. Table 3 presents the comprehensive performances of all models across both datasets.

**Table 3**: **Accuracy percentages of Four Classifiers on BAAD16 and BARD10 Under Two Scenarios**

| Dataset | Stopword Removal | Bangla BERT | XGBoost | SVM | MLP |
|---|---|---|---|---|---|
| **BAAD16** | No | 0.981 | **0.949** | **0.997** | **0.955** |
|  | Yes | **0.983** | 0.918 | 0.995 | 0.953 |
| **BARD10** | No | **0.908** | **0.861** | **0.921** | **0.816** |
|  | Yes | 0.902 | 0.835 | 0.893 | 0.776 |

According to the accuracy percentages presented in Table 3, SVM emerges as the unequivocal victor in both corpora, attaining near perfect accuracy on the long form BAAD16 texts and consistently securing the highest score on the shorter, blog style BARD10. Eliminating stop-words has no consistent benefit: it results in a slight improvement for Bangla BERT on BAAD16, while diminishing or failing to enhance performance for all other model-dataset combinations. The decline is particularly pronounced for XGBoost on BAAD16 and for MLP on BARD10, indicating that Bangla stop-words encompass author-specific style indicators that frequency based and filter based models leverage. The results highlight that, in Bangla authorship attribution, preserving stop-words is crucial for optimizing classification accuracy, while advanced deep models offer minimal advantages given the existing input length and data limitations.

## 4.2 Impact of Stop-word Removal

**Table 4: Impact of Stop-word Removal (ΔF1 is the test set F1 difference between runs without and with stop-words, on the same split and settings, negative values favour retention.)**

| Model | ΔF1 BAAD16 | ΔF1 BARD10 | Interpretation |
|---|---|---|---|
| Bangla BERT | +0.002 | −0.006 | A slight gain was observed on BAAD16 when stop-word tokens were removed, plausibly because tokens were freed within the 512 subword window; for BARD10's blog posts, removal diminished stylistic cues often carried by tokens in the stop-word list. |
| XGBoost | −0.031 | −0.026 | Degradation was observed on both corpora, consistent with reliance on high frequency tokens targeted by the stop-word list. The effect was stronger on BAAD16 and remained material for BARD10's blog posts. |
| SVM | −0.002 | −0.028 | Drops indicate that tokens removed by the stop-word list contribute to discrimination, particularly in BARD10's short blog posts. |
| MLP | −0.002 | −0.040 | Minimal impact was seen on BAAD16 where long inputs fill the 4 000-token frame; on BARD10's short blog posts, removal shortened effective sequences and increased padding, producing a pronounced decline. |

Table 4 indicates that the elimination of stop-words is, at most, neutral and frequently detrimental across both corpora. Bangla BERT achieves a minimal increase of 0.002 F1 on the lengthy, book style BAAD16 texts due to the removal of high frequency tokens, which creates additional capacity inside its 512 token windows, nevertheless, it experiences a decline of 0.006 F1 on the shorter, discourse centric BARD10 postings, where those same tokens convey authorial voice. XGBoost experiences the most significant decrease on BAAD16 due to its TF-IDF features being directly influenced by stop-word frequency, whereas the linear SVM records only a minimal fall, indicating that even sparse models derive stylistic value from stop-words. The MLP is minimally impacted on BAAD16 but fails on BARD10, as the removal of

15% of already brief inputs exacerbates padding and diminishes signal integrity. Aside from a negligible enhancement for BERT on BAAD16, the elimination of stop words consistently fails to improve and frequently diminishes Bangla authorship-attribution accuracy, highlighting the stylistic significance of Bangla stop-words in both datasets. Figure 3 illustrates the confusion matrix for the optimal model on the BARD10 dataset, utilising preprocessing that exclusively removes punctuation. The digits 0 through 9 denote the ten writers in the matrix.

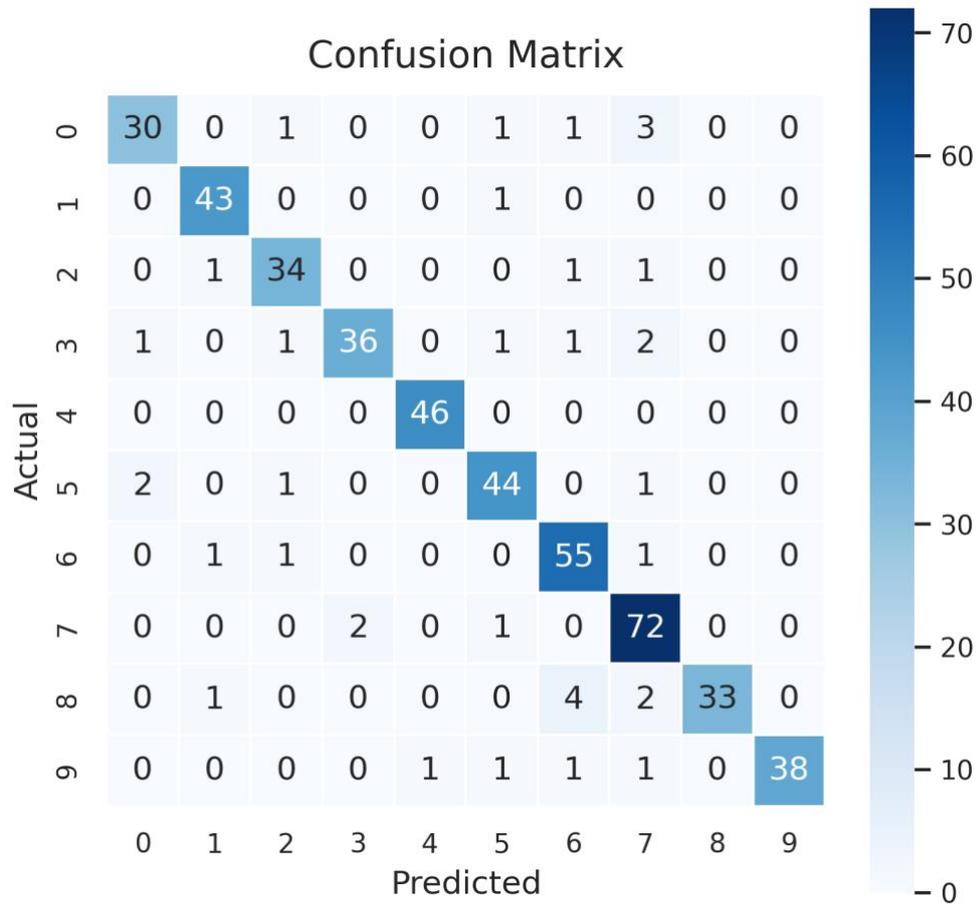

**Figure 3**: Confusion matrix for the SVM classifier on the BARD10 corpus (including stopwords; author labels 0 – 9 as shown in Table 2).

The matrix is highly diagonal, corroborating the elevated macro-F1 score (0.92) reported for the SVM. Author 7 (row 7) is nearly flawlessly identified, with 72 out of 73 texts accurately classified, while authors 6 and 4 exhibit minor misunderstanding as well (55/57 and 46/46 correct, respectively). The majority of errors are singular, isolated swaps; however, a minor cluster of misclassifications is evident between authors 0 and 7, as well as between authors 5 and 8, suggesting stylistic resemblance in those pairs. The limited off diagonal entries demonstrate that TF-IDF features combined with a linear margin effectively encapsulate the unique stop-word and n-gram characteristics of each author in this blog style dataset.

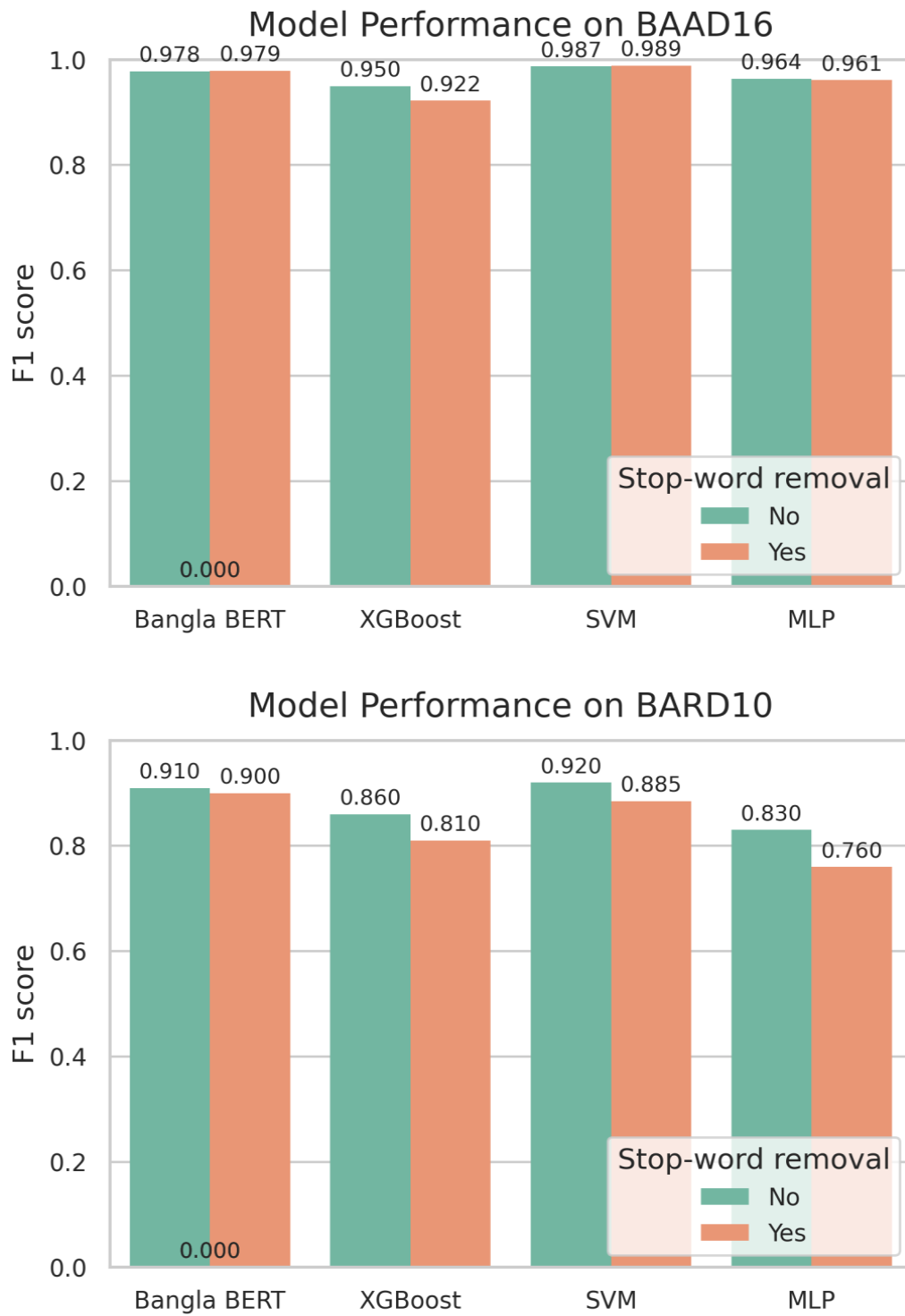

**Figure 4**: Comparison of F1 Scores Across Models and Preprocessing Strategies for BAAD16 and BARD10 Datasets.

Figure 4 shows model performance (F1) on BAAD16 and BARD10 under two different settings, stopwords included and without stopwords. On BAAD16, uniformly high F1 scores were obtained; SVM remained the top performer and XGBoost showed the greatest sensitivity to stop-word removal , whereas Bangla BERT exhibited a slight gain and MLP changed minimally. On BARD10, overall scores were lower, SVM again led but declined when stop-words were removed, as did XGBoost and MLP, while Bangla BERT shifted only modestly. The pattern suggests that eliminating tokens targeted by the stop-word list disproportionately harms models relying on frequency profiles or bag-of-embeddings, particularly on shorter blog texts, whereas the transformer remains comparatively robust.

Having established that indiscriminate stop-word removal is neutral to harmful at system level (Table 4), we next examine which individual stop-words carry authorial signal via a frozen-model ablation that measures $\Delta$-Recall per token.

## 4.3 Token-level Stop-word ablation

To localise which stop-words carry authorial signal, we freeze the trained SVM and, for each candidate stop-word column in the TF-IDF test matrix, set that column to zero and recompute recall per author. $\Delta$-Recall = (baseline recall) − (recall after ablation) is reported, in percentage points (pp). A positive $\Delta$ indicates the token helps retrieve that author (removal hurts); a negative $\Delta$ indicates the token is confusing or noisy (removal helps). This analysis explains why the system level effects in section 4.2 arise by exposing token level contributions to author recognition.

On BARD10, several authors rely heavily on discourse particles and pronouns. Removing a single high impact stem yields pronounced recall losses for those authors, confirming that seemingly "generic" stop-words act as stylistic fingerprints in Bangla web prose. Conversely, a few tokens  exhibit negative $\Delta$ for specific authors: masking them improves recall, suggesting author idiosyncratic overuse that introduces confusion. Together, these results clarify the neutral-to-harmful system level effect of blanket stop-word removal as shown in Table 5. For each author, we list the stop-word whose removal most decreases recall (largest $+\Delta$; harmful to remove) and the stop-word whose removal most increases recall (most negative $\Delta$; helpful to remove).

**Table 5: Per-author extremes (BARD10)**

| Author | Most harmful token (largest +Δ) | 2nd most harmful | Most helpful-to-remove token (most –Δ) | 2nd most helpful-to-remove |
|---|---|---|---|---|
| A0 | — (no token causes ≥ +0.5 pp drop) | — | জন (-5.6 pp) | এট (-2.8 pp) |
| A1 | — | — | — | — |
| A2 | মত (+8.1 pp) | অন (+5.4 pp) | — | — |
| A3 | — | — | — | — |
| A4 | — | — | — | — |
| A5 | অত (+3.5 pp) | এই (+3.5 pp) | লক্ষ (-1.0 pp) | আর (-1.0 pp) |
| A6 | আম (+5.2 pp) | বলল (+3.4 pp) | আপন (-1.7 pp) | বলত (-1.7 pp) |
| A7 | হয় (+1.3 pp) | এত (+1.3 pp) | — | — |
| A8 | এখন (+2.5 pp) | এমন (+2.4 pp) | নয (-2.5 pp) | বহ (-2.5 pp) |
| A9 | কত (+2.4 pp) | হল (+2.4 pp) | — | — |

In BAAD-16, per token Δ values are typically smaller than in BARD10 because chapters are longer and lexical content is richer; nonetheless, a subset of particles (e.g., তর, নও, উপর) still carry measurable authorial signal for specific writers. Table 6 shows per-author extremes of stop-word ablation on BAAD-16.

**Table 6: Per-author extremes (BAAD-16)**

| Author | Most harmful token (+Δ) | Δ (pp) | Most helpful-to-remove (–Δ) | Δ (pp) |
|---|---|---|---|---|
| A0 | একট | +0.57 | — | — |
| A1 | — | — | — | — |
| A2 | সঙ | +0.14 | — | — |
| A3 | — | — | — | — |
| A4 | — | — | বল | −2.78 |
| A5 | উপর | +1.32 | — | — |
| A6 | উপর | +0.50 | একট | −0.50 |
| A7 | — | — | হইত | −0.60 |
| A8 | — | — | — | — |
| A9 | — | — | এক | −0.70 |
| A10 | তথ | +0.48 | হইত | −0.48 |
| A11 | নও | +2.21 | — | — |
| A12 | বলল | +0.64 | হয | −0.32 |
| A13 | — | — | — | — |
| A14 | নও | +1.34 | — | — |
| A15 | তর | +3.33 | — | — |

Figure 5 shows the boxplot of absolute (non-zero) Δ-Recall values compares how strongly individual tokens influence author retrieval in BARD10 and BAAD16. In BARD10, the median absolute Δ-Recall is around 2.5 percentage points and the interquartile range spans roughly 2–2.5 pp; multiple outliers exceed 4 pp, suggesting a few tokens have large beneficial or detrimental effects.

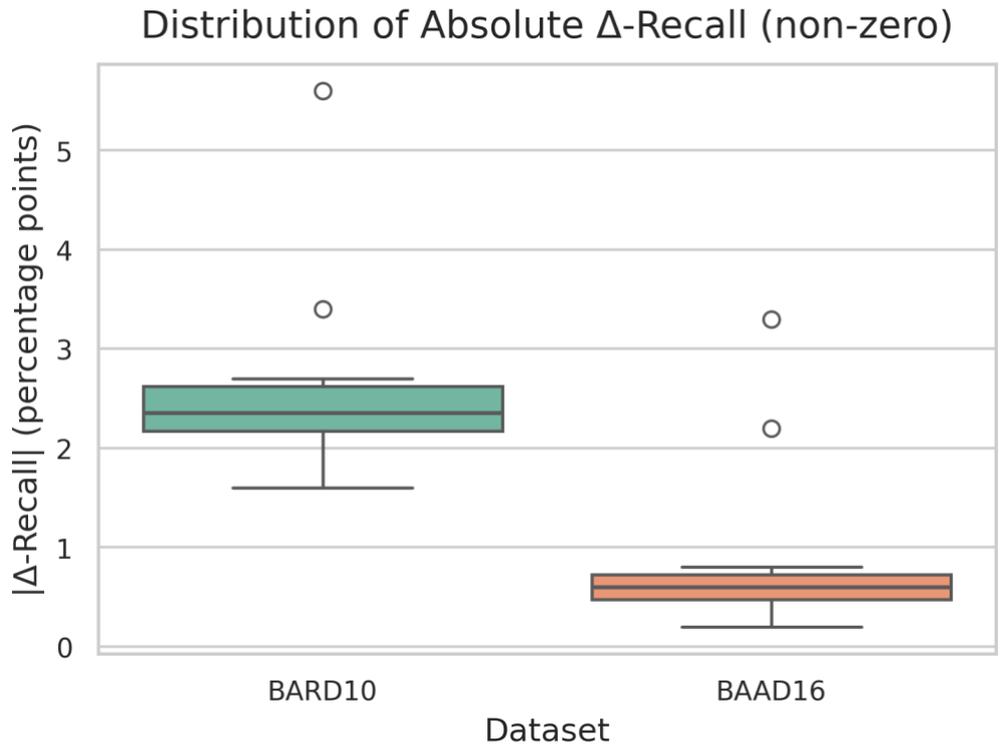

**Figure 5:** Distribution of non-zero absolute Δ-Recall values for BARD10 and BAAD16.

By contrast, BAAD16's distribution is much more compact: most absolute Δ-Recall values lie between 0.2 and 0.8 pp, with a lower median near 0.6 pp and far fewer outliers. These differences indicate that tokens in BARD10 generally have a larger and more variable impact on recall, whereas BAAD16's tokens exert modest, more consistent effects.

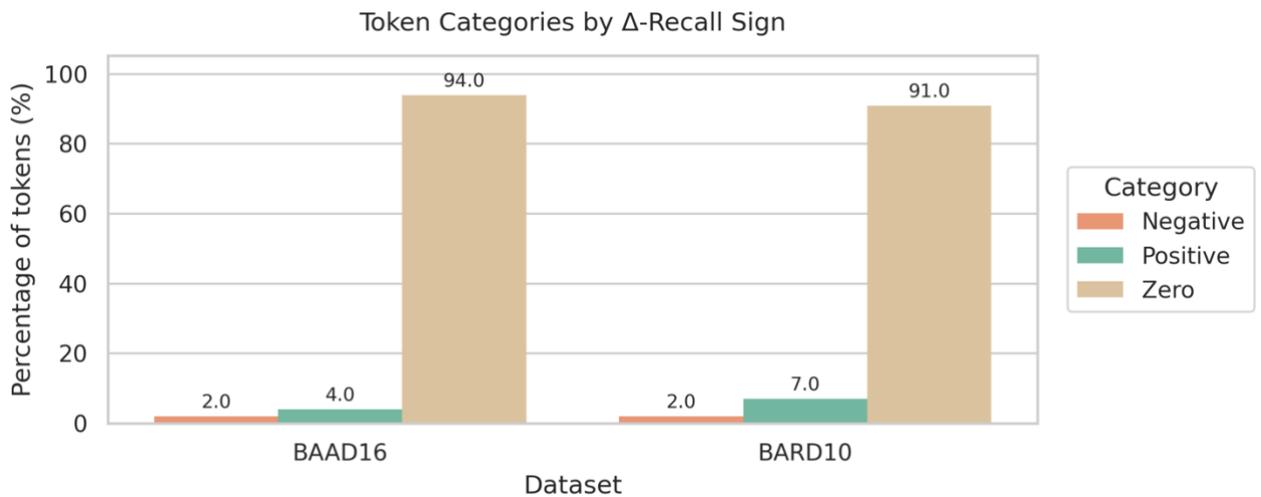

**Figure 6**: Proportion of token–author pairs yielding positive, negative or zero Δ-Recall

The barchart of Figure 6 highlights not just how large the non-zero Δ-Recall values are but also how often they occur. In BARD10 roughly 9 % of token-author pairs have a measurable effect, with about 7.3 % improving recall and 1.9 % harming it; BAAD16 has fewer such pairs, with only 3.9 % positive and 2.2 % negative. Most pairs are zero in both datasets, but the share of non-zero effects is visibly higher in BARD10.

## 5 | Discussion

Experimental results reveal a consistent and distinct trend: traditional TF-IDF-based models, particularly the linear SVM, outperform Bangla BERT and MLP across both BAAD16 and BARD10. Two primary factors contribute to this outcome.

First, input length constraints substantially limit the representational capacity of deep architectures. Bangla BERT discards nearly seventy five percent of a typical BAAD16 chapter when restricted to its 512 token window, while the MLP model inefficiently expends most of its 4,000 token allowance on padding when processing the much shorter BARD10 blog posts. In contrast, TF-IDF features incorporate every available token, irrespective of document length, ensuring that each author's lexical distribution is fully captured.

Second, the models diverge fundamentally in their treatment of high frequency stop-words. In the BARD10 corpus, the proportion of stop-words varies only slightly from 14.3 percent in Noyon to 17.5 percent in Tajerul Islam, yet this seemingly minor three point difference results in thousands of additional stop-word tokens per author. Classical models, such as linear SVM and boosted trees, effectively exploit these variations by assigning high weight TF-IDF n-grams, thereby encoding distinctive stylistic signatures that differentiate authors. Conversely, Bangla BERT's pretraining objective intentionally down weights frequent tokens, causing contextual embeddings of common subwords to converge across authors and consequently suppressing stylistic nuances.

Our Δ-Recall analysis reinforces this interpretation by revealing a clear dichotomy between the two corpora. In BARD10, stop-word removal caused substantial recall declines for authors such as A8, A9, A6, and A2 indicating that their stylistic identity relies heavily on stop-word usage, pronouns, and discourse level connectors. BAAD16, by contrast, exhibited far less sensitivity: only a few authors (A15, A11, and A14) experienced notable declines. This asymmetry stems from corpus composition. BAAD16 comprises canonical literary prose by authors such as Tagore and Nazrul, whose writing is characterized by content rich vocabulary, metaphorical expression, and thematic depth, traits of content stylists. Meanwhile, BARD10 consists of informal, blog style narratives where stylistic fingerprinting often depends on conversational rhythm and first person discourse markers. Removing stop-words in such texts erases crucial rhythmic and pragmatic cues, whereas in BAAD16, authorial distinctiveness persists through content bearing terms.

Together, these findings yield two major insights. First, stop-words act as critical stylistic discriminators in Bangla authorship, particularly within modern, conversational genres, and therefore should not be universally removed during preprocessing. Second, while deep contextual models tend to suppress these lexical subtleties, TF-IDF combined with linear SVM remains remarkably robust in capturing them, establishing a strong, interpretable baseline that remains competitive across both classical and contemporary Bangla corpora.

# 6 | Conclusion

BARD10, a balanced corpus of contemporary Bangla prose, was introduced and used with BAAD16 for a unified evaluation of classical and neural authorship models. Across both datasets, the sparse TF-IDF + linear SVM baseline was found to be most reliable, whereas deeper architectures yielded limited gains under current context length and data constraints. Token level $\Delta$-Recall analyses indicated that Bangla stop-words encode stable authorial signals; consequently, we observe that indiscriminate stop-word removal degrade performances, most notably on BARD10. The BARD10 corpus and reproducible code are released to facilitate future work on selective stop-word handling, long context/domain-adapted transformers, and hybrid systems that combine sparse and contextual representations.

# CRediT authorship contribution statement

# Data availability

BARD10 dataset: https://doi.org/10.5281/zenodo.17572060

# Funding

There was no explicit support for this study from the government, industry and non-profit funding organization. It was self-funded research.